# Words as Difference Makers: How Large Language Models Determine Causal Structure in Text

Wolfgang Pietsch (wolfgang.pietsch@tum.de)

**Abstract:** Because large language models (LLMs) are impressively successful in predicting text, it appears that they must have access to a 'world model' representing causal and definitional structure. However, the dominant formalisms of modern causal inference—Judea Pearl's interventionist approach and the Neyman-Rubin potential outcomes framework—struggle to illuminate how LLMs learn causal structure. I resolve this puzzle by arguing that LLMs employ a specific inductive approach based on a difference-making logic—sometimes called variational induction. I demonstrate how central aspects of this logic are realized during training, where LLMs require enormous amounts of text data from a wide range of contexts to identify difference- and indifference-makers within word sequences. Furthermore, I analyze specific architectural features of LLMs—such as token embeddings and self-attention—to determine their roles in variational induction. The difference-making logic of LLMs fundamentally parallels the experimental method, where causal relations are derived by systematically varying individual circumstances to determine their influence on a phenomenon.

## Content

## 1. Introduction

One of the most remarkable scientific discoveries and one of the most impressive engineering achievements of our time is that multilayered neural networks with hundreds of billions of parameters exhibit sophisticated reasoning-like behaviour. These large language models (LLMs) like ChatGPT do not just recombine the information they are given, but they infer generalized relationships that represent causal and definitional structure in text. Based on these relationships, LLMs generate meaningful word sequences that are nowhere in the text corpora with which these models have been trained. In other words, large language models reason inductively from known to unknown word sequences, and they do so successfully.[1]

Former OpenAI Chief Scientist Ilya Sutskever frames next-word prediction as a form of optimal data compression. To predict the next word, a model is driven to discover the hidden, generative laws of the data, thereby building an internal 'world model' (2022). However, if LLMs map causal structure, their methodology poses an epistemological puzzle. The two dominant frameworks of modern causal inference—Judea Pearl's structural causal models (2009), based on causal graphs and the do-calculus, and the Neyman-Rubin potential outcomes framework (Rubin & Imbens 2015)—both fail to illuminate how large-scale neural networks are capable of inferring causal structure. LLMs neither deploy Pearl's do-calculus nor do they construct Rubin-style counterfactual states.

The present article resolves this puzzle by examining the precise type of induction contemporary LLMs employ. Elsewhere, I have argued that many machine learning algorithms and in particular neural networks implement *variational* or *variative induction*, a type of induction based on a difference-making logic (Pietsch 2021, 2022). Variational induction infers general relationships by systematically varying the circumstances of a phenomenon while observing the resulting impact on the phenomenon. The main method of this type of induction is the *method of difference*.[2] Variational induction has a long and venerable history in expositions of scientific method, dating back at least to medieval scholars like Roger Bacon, Duns Scotus, and William of Ockham (cp. Losee 2001, Ch. 5) and advocated by some of the most influential methodologists since, including Francis Bacon (1620/1994), John Herschel (1851) and John Stuart Mill (1886).

[1] In his pioneering book on "Artificial Intelligence and Scientific Method" (1996), Donald Gillies early on acknowledged that machine learning rehabilitates inductive methods, which had been disregarded for much of the 20th century. Gillies and his son Marco, a computer scientist, have revisited the theses of his book in light of recent advancements in neural networks (2022). Another early discussion about inductive and hypothetical elements in machine learning is the article by David Corfield with two influential figures from AI, Bernhard Schölkopf and Vladimir Vapnik (2009). Also in the 2000s, the epistemologist Jon Williamson has argued that "advances in automated scientific discovery have lent plausibility to inductivist philosophy of science" (2010, 88; see also 2004). More recently, Hykel Hosni and Jürgen Landes, in their edited collection on logics for the methodology of data-driven research, have stressed that "data-intensive and AI-driven science call for a new methodology of formalized inductive reasoning" (2025, v). Nico Formanek (2025) analyses inductive assumptions in machine learning algorithms.

Because large language models essentially are multilayered neural networks specialized for the task of modeling human language, it is plausible to assume that these models employ variational induction as well. The present article is a long and detailed argument for this thesis. Since variational induction resolves causal and definitional structure in text, it constitutes more than mere interpolation or formula fitting. If the training corpora are adequately chosen, this causal structure represents corresponding relationships in the world. Determining which kind of induction contemporary large language models utilize for making predictions can help us understand the inner workings of those models. It can shed light on why these models work at all, what they may be capable of and where principal boundaries lie. Ultimately, understanding the logical underpinnings of LLMs might even provide guidance on how to further improve these models.

## 2. Basics of Large Language Models

### *2.1 Next-Token Prediction*

The core idea of large language models like ChatGPT is astonishingly simple: *next-word* or, more precisely, *next-token prediction*.[3,4] Based on the query asked and the answer to the query generated so far, the subsequent token continuing the answer is predicted. These *tokens* in many cases are just words, but they can also be, for example, parts of words or punctuation marks. The process is repeated until a stop is predicted signifying that the answer is complete. For example, given the query "What is the connection between machine learning and epistemology?" and a token sequence produced so far "Machine learning (ML) and epistemology are connected in several key ways, primarily through their shared focus on the process of knowledge", the model predicts the following continuation of the sequence token by token: "acquisition" "," "justification" "," "and" "understanding".

---

[2] Federica Russo has argued that a rationale of variation, rather than a rationale of regularity is central to causal analyses in the sciences. She develops her view through a detailed study of scientific practices in various fields (2007, 2009). Stefano Canali and Emanuele Ratti discuss applications of machine learning exhibiting a variational approach (2024). To my knowledge, L. Jonathan Cohen first used the term "variative" or "variational induction" for inductive inferences drawing on variation in the relevant evidence, including Mill's methods (e.g.1989).

[3] I developed my understanding of large language models primarily from three sources: Raschka (2025), Wolfram (2023) and with some help of LLMs themselves.

[4] I use the technical term 'token' throughout the article, which could be replaced by 'word' without altering the essence of what is said.

The prediction of each token is based on an enormous neural network model[5] which has been trained on billions of words contained in vast corpora taken from electronic book collections, encyclopedias, and the largest text source of our time, the internet. Correspondingly, contemporary LLMs have billions of parameters that must be optimized in the training process. For example, GPT-3, a predecessor to the models used in ChatGPT, possesses 175 billion parameters (Brown et al. 2020). It is crucial that the training data includes sufficient variation to cover all relevant aspects of the modeled language, while at the same time the model has enough parameters to be able to represent these aspects.

Remarkably, the performance of large language models for next-token prediction, which is measured in terms of a loss function, depends in an almost law-like manner on the number of parameters $N$ of the model, on the size $D$ of the dataset used to train the model, and on the computational resources $C_{min}$ employed for training. Over several orders of magnitude, the relationships almost exactly follow *power laws*, i.e. the test loss $L$ is proportional to $D^{-x}$, $N^{-y}$, and $C_{min}^{-z}$, with different exponents $x$, $y$, $z$.[6] These findings underpin the continued belief among computer scientists that scaling up models improves performance, even at costs amounting to many billions of dollars.

In essence, LLMs are *deep neural networks* with a large number of layers. An input layer receives the values of the input variables, i.e. numbers representing the query token sequence plus the answer sequence generated so far. An output layer of the LLM yields output variables representing probabilities for different tokens that could continue the given token sequence. Each layer of a neural network consists of numerous nodes or neurons, which are connected to nodes in adjacent layers by weighted links. For each link, the strength of the connection is given by a weight parameter $w$, which is adapted during training of the network. Certain non-linear functions, which take the weight parameters $w$ into account, e.g. the ReLU-function[7] $f(x) = \max(0, w_1 * x_1 + w_2 * x_2 + \ldots + b)$, determine how information flows through the neural network from the input to the output layer (cp. Fig. 1).

---

[5] Useful overviews of various philosophical aspects of machine learning are provided in Leonelli (2020/2025), Boge, Grünke & Hillerbrand (2022), Desai et al. (2022), Duran & Pozzi (2025), Trudel et al. (forthcoming). Millière & Buckner (2024) address LLMs specifically.

[6] "These relations hold across eight orders of magnitude in $C_{min}$, six orders of magnitude in $N$, and over two orders of magnitude in $D$. They depend very weakly on model shape and other Transformer hyperparameters (depth, width, number of self-attention heads) […]." (Kaplan et al. 2020, 5; cf. Sec. 4.1 below)

[7] The max operator selects the largest value from the set of numbers in brackets separated by commas. In LLMs, other non-linear activation functions like GeLu or SwiGLU with certain technical advantages are often employed (Raschka 2025, 105).

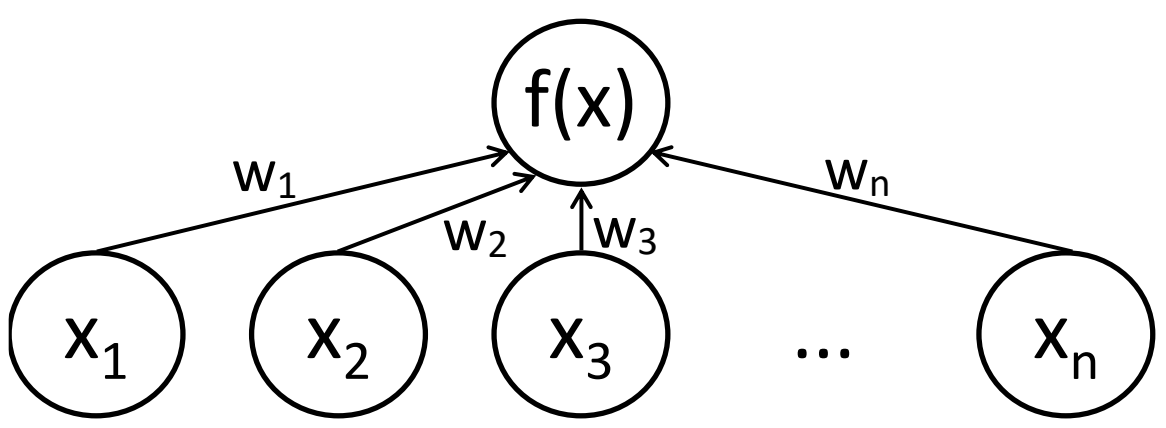

Fig. 1: determination of a value *f(x)* of a node in a subsequent layer given values $x_1, \ldots, x_n$ of nodes in the previous layer

In a simplified reconstruction of a large language model, the input and output take the form of high-dimensional vectors each representing specific tokens in a sequence (see Fig. 2 below). Every vector has the dimension of the vocabulary size corresponding to the number of unique tokens. For example, a language model restricted to simple English has a vocabulary size of approximately 10,000 distinct tokens, GPT-2 of 50,257 (Radford et al. 2019, Sec. 2.2). Some contemporary state-of-the-art models use vocabularies exceeding 200,000 tokens. In the simplified reconstruction, input vectors are so-called *'one-hot' encodings*, where a single vector entry assigned to the represented token is set to one, while all other entries are zero (left side of Fig. 2).[8] The output vector contains multiple non-zero entries representing a probability distribution over the entire vocabulary. According to this probability distribution, the token continuing the input token sequence is predicted (right side of Fig. 2). Notably, the most interesting and most human-like texts result when the model does not always choose the most probable token, but occasionally selects less probable ones (see Sec. 4.3 below).

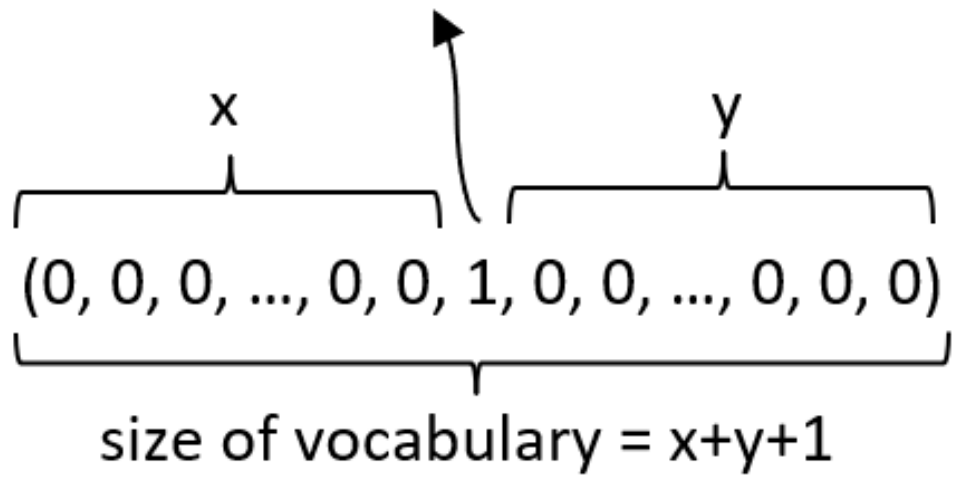

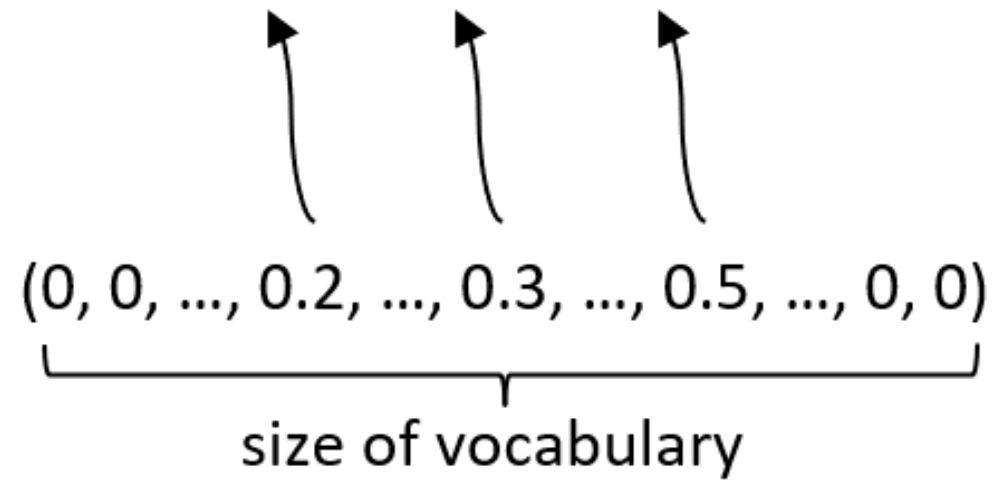

Fig. 2: on the left, a typical 'one-hot' input vector representing a token with ID number *x* (e.g. token ID for the word "induction" is 14355 using the GPT-2 tokenizer); on the right, a typical output vector representing a probability distribution over the entire vocabulary

While large language models are essentially deep neural networks, some characteristics are tailored towards their principal aim: the processing of long sequences of tokens and sequential text generation. The decisive breakthrough in neural network architecture that paved the way for today's successful LLMs came with the introduction of the Transformer model—the capital 'T' in ChatGPT. This model is based on the *attention mechanism* as introduced in the seminal paper "Attention Is All You Need" by researchers at Google (Vaswani et al. 2017). The crucial advantage of the attention mechanism is its ability to handle long-range dependencies in token sequences, which are prevalent in natural language (cf. Sec. 4.3 for a detailed analysis).

[8] To avoid a computationally costly matrix multiplication with a one-hot vector, LLMs employ an 'embedding layer', which uses a token's ID number—as determined by the non-zero entry in the one-hot encoding—to look up the token's embedding vector (see Section 4.2 below).

### *2.2 Pre-Training by Blinding Words*

The 'P' in ChatGPT[9] stands for *pre-trained* describing the initial phase of a training procedure, during which the model learns from massive datasets before being fine-tuned for specific tasks like translation or summarization. In contemporary practice, the fine-tuning has become less important. Modern LLMs are increasingly general-purpose and can handle a wide variety of tasks through prompt engineering, for instance by providing specific examples within the prompt.

LLMs are pre-trained in a *semi-supervised* (or self-supervised) manner. The particulars, with which the model is trained, are ordered token sequences provided in text corpora. Specific tokens within the sequences are *blinded* (or masked). During the training process, the model must predict these missing tokens. The blinding is key, as it allows the model to be trained on massive datasets without human-labeled data, i.e. without manually curating and classifying training data. Any discrepancy between the token predicted by the model and the actual blinded token is used to update the model parameters through an optimization procedure such as *gradient descent*. This process seeks to minimize a loss function which quantifies the difference between the model's predictions and the training data.

For the loss function, LLMs typically employ *cross-entropy loss*, which calculates the difference between two probability functions: the probability distribution predicted by the model and the actual distribution of tokens in the training set. For a single sample, i.e. a single training instance, cross-entropy loss is calculated as follows: $L = -\sum_i p(i) \log q(i)$, where $q(i)$ is the predicted distribution for all tokens $i$ of the entire vocabulary and $p(i)$ is the true or actual distribution of the training instance. The latter is a one-hot distribution, where the probability is one or a hundred percent for the actual blinded token and zero for all other tokens (cf. Section 2.1). The higher the predicted probability for the actual token, the smaller the loss. When the loss function is plotted against the model parameters, including in particular the weights connecting the model's nodes, a high-dimensional loss landscape results (cf. Fig. 3).

[9] 'G' stands for *generative* meaning that the model can generate novel content.

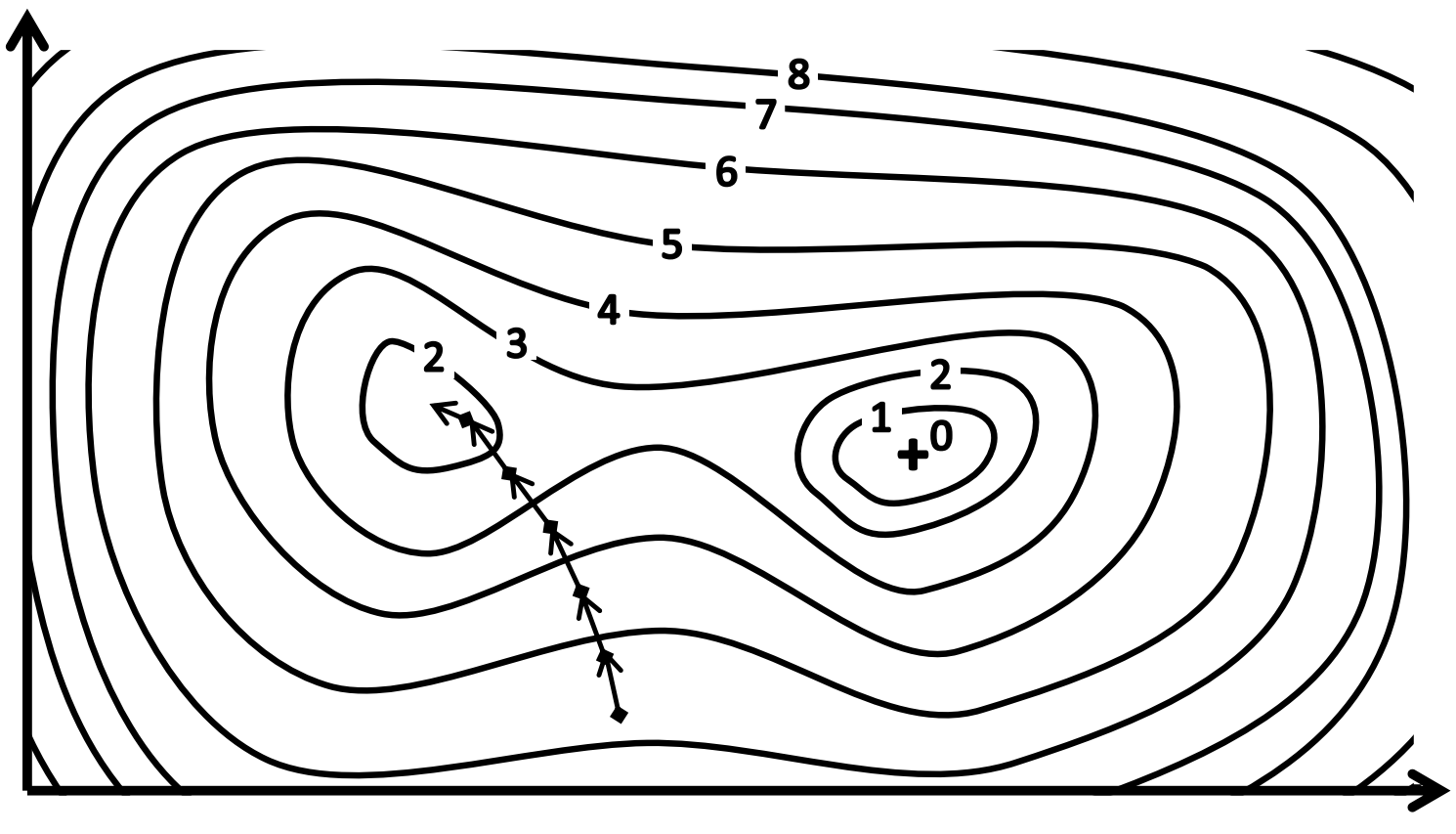

Fig. 3: typical trajectory of a gradient descent to a local minimum in a multidimensional landscape

The current configuration of the model, which is determined by the values of the model parameters at a certain time during training, corresponds to a single point in this loss landscape. Gradient descent, as an optimization procedure, aims to move the configuration towards the landscape's minima, where the loss is lowest and, thus, the deviation of the model's predictions from the training data is minimized. To reach these minima, the slope of the loss landscape is calculated by differentiating the loss function with respect to the model parameters. The parameters are then updated by moving the configuration a predetermined step along the direction of steepest descent. The step size is defined by the learning rate $\alpha$. In LLM training, stochastic optimization procedures such as AdamW with adaptive step sizes are primarily used (Raschka 2025, 148). A typical trajectory is depicted in Figure 3 above. Further training data can be used to evaluate how well the model generalizes, guarding against overfitting and thus ensuring that the model captures meaningful—i.e. in the context of language *semantic*—patterns rather than coincidental relationships.

## 3. Difference / Indifference Making in Large Language Models

### *3.1 Induction in Large Language Models*

In a typical inferential process, the large language model is used to predict a *phenomenon* based on certain *circumstances* (or conditions) of the phenomenon. The circumstances are an ordered sequence of tokens consisting of the query and the part of the answer to the query generated thus far. The predicted phenomenon is the subsequent token following the known sequence.[10]

In the example of Section 2.1, the circumstances consist of the following ordered token sequence: "What is the connection between machine learning and epistemology? - Machine learning (ML) and epistemology are connected in several key ways, primarily through their shared focus on the process of knowledge". This token sequence can be regarded as a single circumstance or, equivalently, as a plurality of circumstances, where each circumstance consists of a token and its relative position: (What, 1), (is, 2), …, (process, n-2), (of, n-1), (knowledge, n). The predicted phenomenon is the next token: (acquisition, n+1). The predicted token, i.e. "acquisition", is then added to the ordered token sequence to form a new circumstance for the subsequent prediction and so on.

[10] Alexander Mussgnug (2022) analyses and problematizes, how machine learning tends to convert various kinds of scientific tasks into prediction problems.

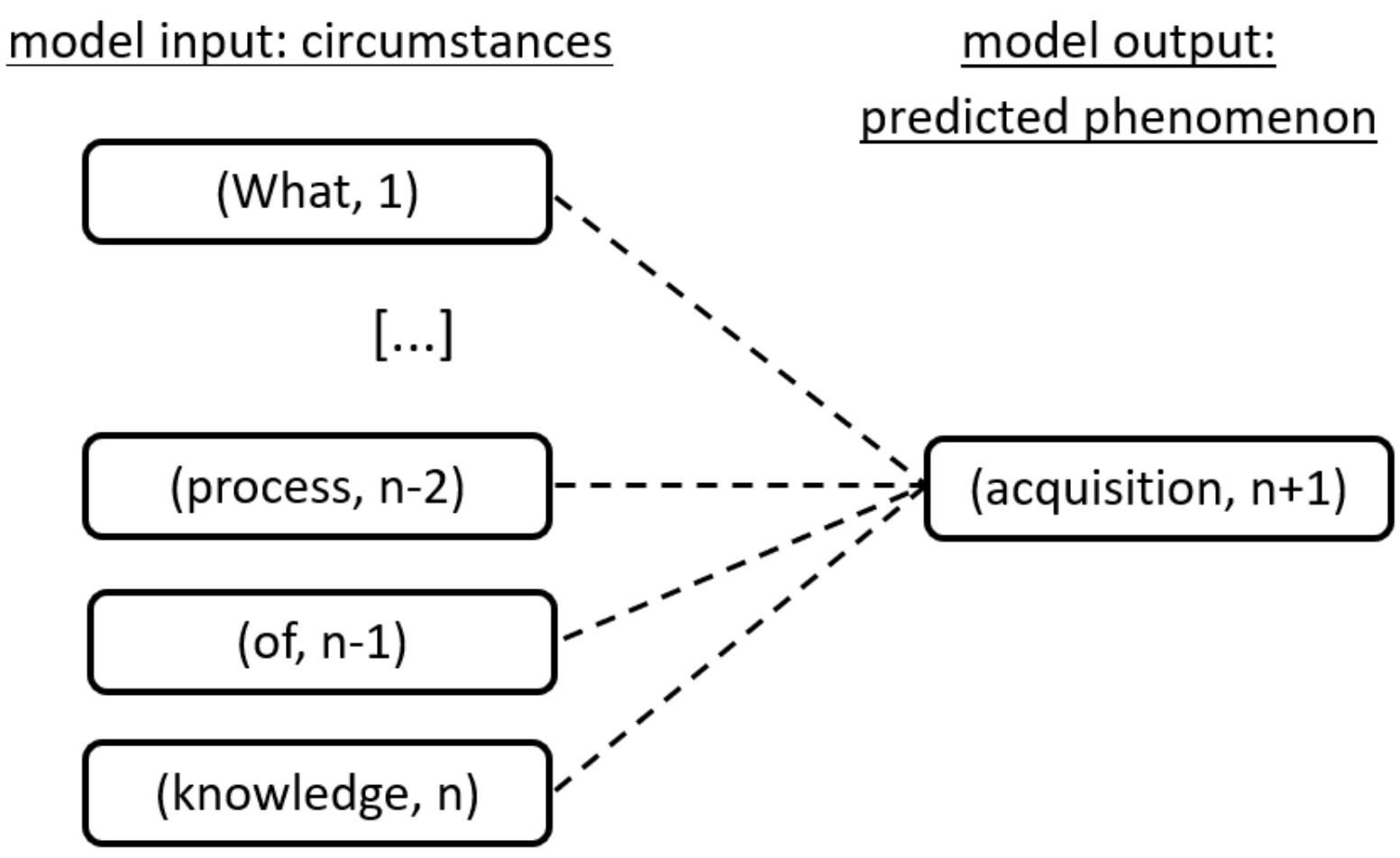

Fig. 4: schema of next-token inferences in LLMs

Two principal paradigms are commonly distinguished in scientific methodology: (i) *inductivism*, which infers general relationships or models from particulars, and (ii) *hypothetico-deductivism*, which posits general relationships or models and tests whether observed particulars are compatible with them. The crucial difference between these paradigms is that hypothetico-deductivism relies on substantial hypothetical assumptions about the modeled subject matter, whereas inductivism does not require such assumptions, deriving its models in an almost mechanical manner from the particulars.

Large language modeling clearly belongs to the inductivist paradigm. As outlined in Section 2.2, LLMs are derived from concrete instances or particulars in the training data using gradient descent. For this procedure, only a small number of assumptions regarding the overall architecture of the model must be presupposed, for example regarding the number of layers in the model or the number of nodes in the different layers (cf. Sec. 4.1 below for more details). Importantly, these assumptions do not concern the modeled subject matter, i.e. human language, but rather they frame the inferential process itself.[11]

[11] For the closely related topic of theory-ladenness of big data practices and machine learning algorithms, see Anderson (2008), Napoletani et al. (2011), Leonelli (2014, 2016), Pietsch (2015), Northcott (2020), Andrews (2025), Termine et al. (2025). A very useful overview and argument in favour of a role for expert knowledge is given in Hansen and Quinon (2023).

Thus, large language models are inductively inferred from the training instances. Accordingly, predictions based on LLMs also constitute inductive inferences. A given token sequence is fed into the model and the subsequent token is determined based on the probability distribution output by the model.[12] Essentially, the token is predicted by comparing the given token sequence with an abstract learned representation of the enormous number of ordered token sequences contained in the training corpora. LLMs rely on induction both for training and for inferences.

Buchholz and Raidl (2025) have argued that artificial neural networks implement a hypothetico-deductive or falsificationist approach, essentially because hypotheses are allegedly refuted during learning and new hypotheses are selected based on some notion of simplicity. However, the optimization procedure of gradient descent is not compatible with the hypothetico-deductivist requirement that scientists (or artificial agents) must rely on their creativity and intuition to choose bold and simple hypotheses which are exposed as much as possible to potential falsifications. Instead, gradient descent implements an incremental learning process, during which the models are only slightly altered in response to bits of new data, instead of being outright discarded.

In summary, large language models implement an inductive approach drawing on a vast number of particulars given by the training corpora and resulting in an equally vast number of generalized ‘laws’ relating truncated token sequences with continuing tokens. In other words, large language models can be interpreted as an aggregation of a massive number of probabilistic laws relating model input with model output, i.e. given token sequences with probability distributions for the subsequent token. Of course, these laws are very complex and strongly context-specific, which distinguishes them from, say, the laws of physics, which are supposed to be simple and to generalize over a large range of contexts.[13,14]

---

[12] Deriving the probability distribution for the subsequent token from the input sequence and the given model is strictly speaking a deductive step. However, the whole process of making predictions based on training sequences is inductive.

[13] Due to the enormous complexity and the resulting opacity of machine-learning models, I am sympathetic to Vladimir Vapnik’s view that the inferential process can be interpreted as a more or less direct inference from training particulars to test particulars—a process he calls *transduction*: “When solving a problem of interest, do not solve a more general problem as an intermediate step. Try to get the answer that you really need but not a more general one.” (Vapnik 2006, 477)

[14] Kuhlmann (2011) introduces a useful distinction between *compositional* and *dynamic complexity*. The compositional complexity of many phenomena modelled through machine learning lies at the origin of ongoing debates regarding explanation and understanding provided by AI models (Knüsel & Baumberger 2020; Beisbart & Räz 2022; Sullivan 2022; Meskhidze 2023; Tamir & Shech 2023; Räz & Beisbart 2024; Beisbart 2025; Greif 2025). We are still in the process of finding out to what extent machine learning and data-intensive

### *3.2 Difference Making and Variational Induction*

Having established that large language models employ an inductive approach raises the question what kind of induction they implement? Several types of inductive inferences are known from debates on scientific method and everyday reasoning: e.g. *enumerative induction* or *eliminative induction*. Elsewhere, I have argued that, from an epistemological point of view, the most successful machine learning methods—including neural networks—rely on a *difference making logic* or *variational induction*[15] (Pietsch 2021, Ch. 4.2; see also 2022, Ch. 4). In the following, I will show that this type of induction also underlies the inferential approach used in large language modeling.

Variational induction has a distinguished history in analyses of scientific methodology. Core methods like the method of difference or the method of agreement can be traced back to Francis Bacon's *tables of discovery* (1621) and even further to medieval predecessors like Roger Bacon, Duns Scotus, and William of Ockham (cp. Losee 2001, Ch. 5). The best-known—if somewhat flawed—account is due to John Stuart Mill (1886). Other notable proponents include Bernard Bolzano (1837/1972; see also Cohen 1990), John Herschel (1851), John Maynard Keynes (1921), Georg von Wright (1951), John L. Mackie (1980, appendix), Brian Skyrms (1966), and Baumgartner and Graßhoff (2004). The approaches of these authors differ substantially. To make my argument regarding the role of variational induction in LLMs, I will in the following rely on the framework developed in Pietsch (2016b; 2022, chs. 5, 6; 2026b).[16]

To make inferences, variational induction draws on evidence that concerns systematic *variation* of circumstances and the impact this variation has on a phenomenon. The fundamental idea of variational induction is that an inductive inference is more reliable the more varied the relevant evidence is. In other words, confirmation of a generalization increases with the *variety* of observed instances. By contrast, *enumerative induction*—which is widely held to be the quintessential, if fallible inductive method—focuses on regularity or repetition rather than variation. Essentially, in enumerative induction, confirmation is thought to increase with the *number* of observed instances.

The principal method of variational induction is the *method of difference*, by which the (causal) *relevance* of a circumstance to the phenomenon under investigation can be determined. Following Mill, the method of difference and related methods like the method of agreement or the method of concomitant variation are often referred to as *eliminative induction*. However, this terminology is unfortunate, since, at least in modern literature, *eliminative induction* also refers to another inferential approach, in which hypotheses are

---

approaches can help us predict and interact with a complex, dappled and fragile world (Cartwright 1999, Northcott 2025).

[15] According to the terminology of this article, variational induction includes any account of induction that implements a difference making logic.

[16] A valuable critical assessment is given in Galli (2023), to which I reply in Pietsch (2026a).

eliminated from an exhaustive set until only the correct hypothesis remains. For inferential methods based on a difference-making logic, I therefore prefer the term 'variational induction' to differentiate it from the mentioned type of eliminative induction and also because methods such as the method of difference are not inherently eliminative in nature.

In a typical reconstruction of the *method of difference*, two instances are compared that differ only in a single, potentially relevant circumstance. If the circumstance and the examined phenomenon are both present in one of the instances, but both absent in the other instance, then the circumstance has causal relevance to the phenomenon (e.g. Mill 1886, 256; Herschel 1851, 154-155).

As an example, consider a set of light switches, which constitute the circumstances, and a set of ceiling lights, which constitute the examined phenomena. By individually operating the switches and observing whether the lights are on or off, causal relationships between the switches and the respective lights can be established. Here, causal relationships are essentially those relationships, which allow for successful manipulation of the phenomenon. In a simple case, it can be determined that a certain switch controls a specific light.

A counterpart to the method of difference, a variant of the method of agreement which has been called the *strict method of agreement*, can be used to infer the (causal) *irrelevance* of a circumstance to the examined phenomenon: If a change in a circumstance between two instances has no impact on the examined phenomenon, then, under certain premises, that circumstance is causally irrelevant to the phenomenon. For example, by this method one can determine that a certain switch is irrelevant to a given ceiling light by comparing two instances for which the state of the switch changes, but the state of the light does not, i.e. the light stays on or off.

A crucial premise for both methods is the *homogeneity condition*, which essentially requires that the considered instances agree in all potentially relevant circumstances except for those circumstances whose impact on the phenomenon under investigation is explicitly examined.[17] Indeed, in order to find out which switch turns on which light, one must take great care to flip only one switch at a time while leaving the states of all other switches—constituting the other potentially relevant circumstances—unchanged. Fortunately, we usually have robust intuitions as to which circumstances are potentially relevant and which are irrelevant. Thus, when applying the method of difference or the strict method of agreement, the set of potentially relevant circumstances has often been narrowed down to a manageable number of accessible and manipulable variables.

While homogeneity requires that potentially relevant circumstances whose impact is not explicitly considered remain constant, circumstances that are known to be irrelevant to the examined phenomenon may vary from instance to instance. In the above example, an oven switch or a refrigerator switch are irrelevant circumstances that need not be held constant, since such switches are known to be unrelated to the ceiling lights. As a result, any inferred causal relationship holds only *relative to a homogeneous background*, which is defined by potentially relevant circumstances that are held fixed and by irrelevant circumstances that are allowed to vary (Pietsch 2022, sec. 5.2).

[17] For attempts at a precise formulation of the homogeneity condition, see Baumgartner & Graßhoff 2004, sec. X; Pietsch 2022, sec. 5.2.1.

Unfortunately, in the limited space available, it is not possible to introduce a refined account of variational induction that can address all the problems and seeming contradictions arising from the foregoing sketch. While the methods discussed play a crucial role in inductive reasoning, the challenge is to precisely state the premises under which such inferences are warranted. For attempts in this direction, see, for example, Herschel (1851), Mill (1886), Mackie (1980, appendix), Skyrms (1966), Baumgartner & Graßhoff (2004), and Pietsch (2022).

Clearly, the method of difference for determining causal relevance and its twin method for determining causal irrelevance are successfully applied across the sciences as well as in everyday life. Large parts of so-called *exploratory experimentation*—i.e. experiments that are conducted without having substantial theoretical preconceptions—rely on these methods.[18] In experimental contexts with little knowledge about the phenomena of interest, variables are generally varied one at a time, while all other variables are held constant, and the impact of the variation is examined. The example of the light switches illustrates such an exploratory approach. How else would one determine how to reliably turn on the lights in an unfamiliar living room without theoretical knowledge of the cables in the walls or an empirically grounded hypothesis about which switches are linked with which lights?

*Randomized controlled trials*, which constitute the evidential gold standard in many sciences dealing with complex phenomena, implement a statistical variant of the method of difference. For example, in a typical medical setup, one group of patients is given a medicine, while another comparable group receives a placebo. If, on average, the first group fares better than the second, this proves that the medicine had an effect. Of course, one has to be very careful that the impact of any further potentially relevant circumstance is either constant for all patients or at least equal for both groups. In other words, homogeneity has to be ensured.

In the following, I will argue that the training procedure of LLMs can be mapped on variational induction. The result is reassuring since it shows that LLMs rely on the—at least in my view—only inductive method that in the past has consistently proven effective across the sciences. Nothing new happens under the sun.

### *3.3 Variational Induction in LLMs*

In order to substantiate the argument that large language models rely on variational induction, I will now list core features, which are characteristic of this inferential method and which distinguish it from other types of induction. For each feature, I will briefly indicate how it is implemented in large language models.

[18] Exploratory experimentation is prevalent in the engineering sciences, where theoretical knowledge is often scarce (Kuhn 2023, sec. 2.2.2).

*(1) Boolean Combinations of Circumstances*

The basic framing of a problem of variational induction is that evidence is given in terms of individual instances. In each *instance*, the state of a *phenomenon of interest* is paired with a Boolean expression of its associated *circumstances*.[19] Such a Boolean expression combines the states of various circumstances of the phenomenon using the logical operators AND, OR, and NOT. For example, the phenomenon could be a specific ceiling light, while the circumstances include certain wall switches. An instance then consists in an observation of whether the ceiling light is on or off together with the corresponding states of the wall switches.

In large language modeling, *instances* are given in terms of truncated token sequences and corresponding tokens continuing these sequences. Every truncated sequence is split into single tokens and every token together with its position in the sequence can be considered a separate *circumstance*. The *phenomenon of interest* corresponds to the subsequent token continuing the sequence (cp. Section 3.1). If different token sequences result in the same predicted token, this can be expressed by using the operators AND, OR, and NOT combining the individual tokens of the sequences. Clearly, the evidence used for training LLMs has exactly the form required for variational induction.

*(2) Variational Rationale[20]*

Both the method of difference and its counterpart, the strict method of agreement, require instances, where the circumstances and the phenomenon of interest are systematically varied. The more variation, the better, since in this way, the relevance or irrelevance of many individual circumstances to different phenomena can be mapped. The method of difference requires *positive instances*, where the phenomenon of interest is present, as well as *negative instances*, where it is absent. In the context of LLMs, negative instances are those, where tokens other than a specific token of interest follow.

When looking at the type of evidence, with which the most powerful LLMs are trained, variety of evidence rather than multiplicity and repetition is crucial for successful model building. This is well illustrated, for example, by the following quote from an article by OpenAI scientists introducing the GPT-2 model: "Most prior work trained language models on a single domain of text, such as [news articles, Wikipedia, or fiction books]. Our approach motivates building *as large and diverse a dataset as possible* in order to collect natural language demonstrations of tasks in *as varied of domains and contexts as possible*." (Radford et al. 2019, Sec. 2.1; emphases added) GPT-2 was trained on 40 GB of text (ibid.). The successor model GPT-3 was already trained on a 570 GB corpus roughly equivalent to 400 billion tokens (Brown et al. 2020, Sec. 2.2)—corresponding to about 400,000 books the size of the Bible.[21]

[19] The simplest version of the method of difference requires binary variables. However, the framework can be generalized in a straightforward manner to discrete variables having more than two values and even to continuous variables (Pietsch 2022, Sec. 5.3.4).
[20] Federica Russo has argued that a *variational rationale* underlies many methodological approaches in the sciences (e.g. 2007, 2009), a viewpoint that is further corroborated by the present analysis of LLMs.

During pre-training (cf. Sec. 2.2), large language models process the training data in batches, each consisting of multiple fixed-length token sequences. The massive text corpora used for training are typically shuffled at a very large scale. This ensures that a single batch doesn't contain only "Wikipedia articles" or only "Python code," but rather a *representative mix* of the entire dataset preventing the model from forgetting one type of data while learning another. Thus, variation is key on all levels, with respect to the training data as a whole, but also regarding the batches, i.e. the chunks of data which the model is fed during training for updating the model parameters (e.g. Devlin et al. 2018, Sec. A.2).

In a standard procedure called *deduplication*, repetitions of the same sentences are generally deleted from the training set, which according to Google researcher Katherine Lee et al. (2022) increases the speed of model convergence and the overall quality of the resulting models. Apparently, an enumerative approach focusing on multiplicity and repetition of evidence is incompatible with typical procedures of preparing evidence used for training LLMs. Sometimes high-quality data is used several times for training to give it more weight, but this measure is certainly compatible with the variational stance.

Of course, the variation required in the training data is always relative to the complexity of the modeled phenomena. More precisely, for variational induction, the instances with which the model is trained should cover all or most relevant combinations of circumstances and phenomena (Pietsch 2015). Certainly, there are many more admissible token sequences, i.e. a lot more meaningful texts exist, than can be found in the training corpora. However, since LLMs analyze on a semantical rather than a syntactical level, not all these admissible token sequences are necessary for model building (see Sec. 4.2 below on the concept of embeddings). In the end, whether sufficient data was available for training can only be evaluated in hindsight by evaluating whether the model provides adequate predictions, i.e. answers correctly to user queries. Conversely, hallucinations, where the model invents content, occur when the data on a given topic was insufficient.

---

[21] Big data or data-intensive practices and machine learning are closely interrelated topics, since to analyse the former the latter is often required (for philosophical perspectives on big data, see for example Leonelli 2014, 2016, 2020/2025, Canali 2016, Norvig 2017, Haig 2020, Hübner, Frisch & Feest 2021).

*(3) Relevance and Irrelevance of Circumstances to a Phenomenon*

Variational induction can determine—by means of the methods of difference and agreement—whether changing certain circumstances has an effect on the examined phenomenon—i.e. whether these circumstances are relevant or irrelevant to the phenomenon. In this way, variational induction can identify necessary relationships between circumstances and phenomena. The necessity can result, for example, from a causal or a definitional connection.

When large language models are used for inference, the input consists of a token sequence comprising the user query and the answer generated thus far. The input token sequence is then mapped to a corresponding sequence of embedding vectors—each having a dimension significantly smaller than the vocabulary size (cf. Secs. 2.1, 4.2). The entries of these embedding vectors constitute the nodes or neurons of an initial layer of the LLM. Starting from the initial layer, the model processes the sequence of embedding vectors through successive layers, wherein weights for the connections between layers have been trained to capture complex linguistic and empirical relationships. Finally, the model outputs a vector of the same dimension as the vocabulary size, where each entry of the output vector gives the probability that a token associated with this entry continues the input sequence (cp. Fig. 2). Most of these probabilities will be near zero with only a small number of tokens having probabilities significantly larger than zero.

Thus, when text is inferred by an LLM, the *relevance* or *irrelevance* of certain circumstances—i.e. of specific tokens in the input sequence—to the prediction of a subsequent token is modeled by the weights determining the strengths of the connections between the nodes in different layers. By sending the input token sequence through the different layers of the LLM, the weights together with respective activation functions determine which tokens are taken into account for the prediction. Broadly speaking, irrelevance can be modelled by weights 'zero' or 'near-zero' and relevance by weights that differ substantially from 'zero' (cp. Fig. 1).

Before training the LLM, the weights of the connections between subsequent layers are typically initialized to small random values. Initializing to zero would cause a symmetry problem, where the nodes in the network cannot be distinguished during learning. The initialization to small, similar weights shows that the training of LLMs starts from the assumption that the tokens in given input sequences are all of similar relevance for the subsequent token. Thus, the distinction between relevant and irrelevant tokens has to be learned from instances during model training.

Relevance and irrelevance relations derived by variational induction typically concern Boolean combinations of circumstances involving the operators AND, OR, and NOT. For example, in the sentence "the English word 'dog' translates to the German 'Hund'", both the tokens "dog" AND "German" are relevant to "Hund." These Boolean operators can be modeled by multilayered networks employing non-linear activation functions such as the widely used function ReLu $f(x) = \max(0, w_1{*}x_1 + w_2{*}x_2 + \ldots + b)$ (cp. Section 2.1).

For example, let $f(x_1, x_2) = \max(0, w_1x_1 + w_2x_2 + b)$ be the activation function with weights $w_1$, $w_2$ and bias $b$; $x_1 = 1$ represents the presence of a first token $X_1$ in the input sequence, $x_1 = 0$ its absence; $x_2 = 1$ represents the presence of a second token $X_2$, $x_2 = 0$ its absence. Then, $X_1$ AND $X_2$ can be mapped by choosing appropriate parameters, e.g. $b = -1$, $w_1 = w_2 = 1$, which yields: $f = 1$ only for $x_1 = x_2 = 1$ and $f = 0$ otherwise.

$X_1$ OR $X_2$ can be mapped using $b = 0$ and $w_1 = w_2 = 1$ resulting in: $f = 0$ for $x_1 = x_2 = 0$; $f = 1$ for $x_1 = 1$, $x_2 = 0$ and for $x_1 = 0$, $x_2 = 1$; and $f = 2$ for $x_1 = x_2 = 1$. By subtracting $X_1$ AND $X_2$, e.g. in a further network layer, $X_1$ OR $X_2$ can be represented: $f^* = 0$ for $x_1 = x_2 = 0$; and $f^* = 1$ otherwise. By twice subtracting $X_1$ AND $X_2$ in the further network layer, $X_1$ XOR $X_2$ can be mapped: $f^* = 1$ for $x_1 = 1$, $x_2 = 0$ and for $x_1 = 0$, $x_2 = 1$; and $f^* = 0$ otherwise.

Given the activation function $f(x_1) = \max(0, w_1x_1 + b)$, the expression NOT $X_1$ can for example be mapped by $w_1 = -1$ and $b = 1$ yielding: $f = 1$ for $x_1 = 0$; and $f = 0$ for $x_1 = 0$. Some operators can be represented only by using hidden layers between the input layer and the output layer. Hidden layers can also be used to construe complex functions consisting in Boolean combinations of a large number of circumstances.

Of course, these brief derivations only prove that modeling Boolean operators in terms of non-linear activation functions and hidden layers is feasible. The real weights and biases in actual LLMs almost always differ from whole numbers like zero or plus/minus one. Because Boolean operators are a special case, the architecture of neural networks can be considered as *a generalization of the deterministic binary Boolean logic of conventional variational induction to continuous probability distributions over concepts represented by the nodes in the different layers of a neural network*. For example, such probability distributions are derivable by the softmax function introduced in Sec. 4.3.

Due to the input-output structure, every large language model in principle corresponds to an *aggregation of a vast number of probabilistic laws, where each law relates Boolean combinations of ordered input tokens with output probability vectors*. These laws encode information about which input tokens at their respective positions are relevant or irrelevant to which predicted token. Such probabilistic laws correspond exactly to the type of relevance/irrelevance relations that are derivable through variational induction.

*(4) Hierachies of Laws and Concepts*

Relevance and irrelevance can be analyzed by means of variational induction at different levels of resolution with regards to higher-level and lower-level concepts and laws (Pietsch 2022, sec. 5.3). In conventional variational induction, these hierarchies are expressed in terms of Boolean combinations of circumstances. For example, if a higher-level concept $X = (X_1$ AND $X_2)$ OR $X_3$ is causally relevant, then the lower-level individual circumstances $X_1$, $X_2$, $X_3$ will also have causal relevance under certain premises.

In large language models—as in other neural networks—hierarchies are encoded in terms of different layers of a neural network with the nodes in the intermediate layers representing hidden concepts. The influence of nodes in a previous layer is weighted to determine the values of nodes in the subsequent layer (cf. Fig. 1). As discussed under the previous item (3), these weighted links between nodes in subsequent layers rarely represent simple Boolean operations. Instead, the weighted connections and the continuous numbers assigned to nodes or neurons can be interpreted as a probabilistic generalization of the Boolean hierarchies of conventional variational induction.[22]

In contrast to other applications of neural networks, e.g. image recognition, where the input consists of individual pixels of an image and the output of high-level concepts represented by the image, the input and output of LLMs are on the same hierarchical level—both concern individual tokens. However, different hierarchies of representation are still realized in the hidden layers of LLMs, e.g. regarding grammatical, definitional, or empirical relationships at various levels of abstraction and generalization. By relating these different levels to each other, deductive inferences can be represented by LLMs.[23]

*(5) Logic of Difference / Indifference Making*

Variational induction as introduced in Section 3.2 identifies difference and indifference makers among the circumstances of a phenomenon. Obviously, large language models do not directly apply the method of difference and the strict method of agreement. However, as will be shown in the following, procedures like gradient descent, which optimize the parameters of the LLM should be understood in terms of *gradual learning of difference and indifference makers among the circumstances*. This is because, given sufficient variation in the evidence, a model that correctly identifies difference makers and indifference makers will have a smaller loss function than a model that gets the classification wrong.

A simple argument explains why gradient descent will eventually determine difference makers. If the evidence encompasses sufficient variation, it will include both positive and negative instances, i.e. instances where the difference making circumstance or circumstances and thus also the examined phenomenon are present and instances where this or these circumstances are absent and, consequently, the examined phenomenon is absent as well. Ideally, the evidence should furthermore include instances, where all other circumstances apart from the difference making circumstance(s) vary as much as possible. Now, only if the difference maker is identified by the model, both negative and positive instances will be correctly classified—resulting in an overall smaller loss function. Therefore, a model optimized by gradient descent, if it reaches some measure of success, should get at least some of the difference makers in the modeled language right.

[22] As discussed in Section 4.1, according to the Transformer architecture of LLMs, each layer consists of a self-attention module combined with a feed-forward module. Thus, the connections between subsequent layers are more complex than simple weighted links. However, this does not change the overall assessment that these layers can realize hierarchical modelling.

[23] To achieve the real rigor of elaborate mathematical or logical proofs, it appears that LLMs need external help from a proof assistant (Hubert et al. 2025).

A similar argument can be made with regards to indifference making. For example, if an indifference maker is falsely identified as a difference maker, either a corresponding positive or negative instance of the phenomenon will be misclassified—resulting in a larger loss function. Of course, this argument again presupposes that such instances are included in the training set.

Finally, if the background or context, with respect to which difference making or indifference making holds, is not precisely identified by the LLM, this will also result in a larger loss function due to misclassification of training instances having this background.

For example, consider the following four sentences in a training set as basis for an inductive inference: "the English word 'dog' translates to the German 'Hund'"; "the English token 'dog' translates to the German 'Hund'"; "the English word 'dog' translates to the French 'chien'"; "the English word 'cat' translates to the German 'Katze'". For training purposes, the last word in each of these sentences shall be blinded. A model that does not recognize that replacing "German" by "French" or "dog" by "cat" is relevant to whether "Hund" is predicted or not, classifies at least one of the instances wrong and will therefore have a larger loss function. Equally, a model that does not recognize that replacing "word" by "token" has *no* impact on whether "Hund" is predicted, also results in a larger loss function.

This is not to say that neural networks will always identify the correct difference and indifference makers. Rather, there are all kinds of ways in which optimization algorithms like gradient descent might err. For example, it may not find the overall minimum yielding the best classification but may get stuck in a local minimum. Also, there may not be sufficient variation in the training data to distinguish between alternative relevance-irrelevance relations.

A further difficulty for representing difference and indifference making with LLMs is that these models are probabilistic in the sense that the output vector provides probabilities for various tokens which could all continue the input token sequence. Therefore, difference and indifference making need to be understood in a way that changing certain circumstances does or does not have an impact on the *probability distribution* for the subsequent token. These probabilities can be interpreted as limiting relative frequencies given relevant circumstances, i.e. given certain values for the difference making tokens in the input sequence. Space is too limited here to address any additional conceptual issues arising from this probabilistic nature.

Remarkably, using *cross-entropy loss*,[24] as opposed to many other choices of loss function, ensures that *the predicted probability distribution converges towards the corresponding relative frequencies in the training data set*. Let $y$ be the probability of a given token and $1–y$

[24] cf. Sec. 2.2

the probability of an exhaustive set of alternative tokens, both determined by relative frequencies in the *training set*. Let $x$ be the probability of this token and $1–x$ the probability of the set of alternative tokens as predicted by the *model*. Calculating cross-entropy loss for all $N$ instances of the training set, in which either the given token or the alternative set of tokens is present together with the determining circumstances, yields: $-N y \ln x - N (1-y) \ln (1-x)$ (cp. Sec. 2.2). To find a minimum or maximum of this function, we differentiate with respect to the variable $x$ and set equal to zero: $0 = -y / x + (1-y) / (1-x)$. A little arithmetic shows that the equation holds, i.e. the loss function is extremal—and, indeed, minimal—only for $y = x$, i.e. when *the probability predicted by the model is equal to the relative frequency in the training data*. As an example, a well-trained model will use a word more often than its synonym, if the word is more frequently used in the training data.[25] Generally speaking, during the training process, the probabilities predicted by the model will approximate the relative frequencies in the training data and, by correctly representing the probability distributions, the model allows for a probabilistic analysis of difference and indifference making. Due to the above result, a well-trained LLM maps how probability distributions over predicted tokens vary when difference or indifference makers in the input sequence are changed.

As an example, consider a scenario where circumstance A (e.g., rain) causally determines phenomenon C (e.g., a wet street). A proxy variable B (e.g., people wearing raincoats) correlates with C in 95 percent of instances. An optimization procedure aiming to reliably predict C based on A or B should eventually conclude that B must be ignored. Even a slight reliance on B—compared to relying solely on A—would lead to misclassifications. By contrast, if the prediction is based only on A, all instances of C will be correctly classified.

In summary, the difference making logic underlying the training process guards against incorporating spurious or accidental relationships in the large language model as long as there are sufficient training instances. Learning difference and indifference makers rather than merely interpolating ensures that the model generalizes well to yet unknown instances. Of course, this can always be verified by means of a validation set of data that was not used for training or by means of novel queries posed by human users, who find the responses of an LLM helpful. Also, in the (pre-)training of LLMs, novel instances are constantly introduced (cp. item (2) above). Thus, the model is continuously validated, which guards against overfitting during the model building process.

*(6) Simplicity and Underdetermination*

Serena Galli (2023) has rightly stressed that models inferred by variational induction exhibit underdetermination, although, as argued in Pietsch (2026a), this underdetermination importantly does not concern relationships of manipulability. A simple example of such underdetermination is when, in the training data, two or more tokens always co-occur in the same ordering. As long as there are no instances, in which this structure is further resolved, then relevance can always be attributed to either one or both of the tokens.

[25] When employing the LLM for inference, this relationship can be modified by means of a temperature parameter (cf. Sec. 4.3).

For instance, given only the following two token sequences: “the English word ‘dog’ translates to the German ‘Hund’”; “the English token ‘cat’ translates to the French ‘chat’”, there is insufficient information for determining the difference-making circumstances for the change from “Hund” to “chat”—because it can be due to the variation of any of the following three variables or a combination of them: “word” to “token”, “dog” to “cat”, “German” to “French”. A model might even assume that different combinations of circumstances are relevant in different contexts. However, such arbitrarily complex models, postulating structure for which there is no evidence, often generalize poorly. To address this issue, some mechanism is required that selects a simple model, while preserving all relevant information.

Neural networks have architectural features, which force simplicity in modeling. Most importantly, when the number of nodes or neurons in an intermediate layer is smaller than in previous layers, LLMs are forced to compress information (this mechanism underlies the concept of embedding discussed in Sec. 4.2 below). The model must choose a simple alternative, which in the example above could be the one, where the three words together are considered as a single expression to which relevance is attributed. Of course, the ratio between the numbers of nodes in subsequent layers must be diligently chosen so that no relevant information is lost.

Simplicity in model building is sometimes held to be an indicator for a hypothetico-deductive approach (Buchholz & Raidl 2025). However, an algorithmic criterion for choosing simple laws and models, such as the change in number of nodes per layer described above, is certainly compatible with inductive methods (cf. Sterkenburg 2025). Such mechanisms are well suited to resolve the underdetermination present in variational induction.

*(7) Homogeneity*

*Necessary* relationships between circumstances and a phenomenon can be inferred by variational induction only when the *homogeneity condition* is met. This condition essentially requires that only irrelevant circumstances are allowed to vary between instances which are compared by the methods of difference and agreement—except, of course, for those circumstances whose impact is explicitly examined (cp. Section 3.2). Inferring necessary relationships is essential for determining the meaning of texts.

In large language modeling, homogeneity in *inference* and homogeneity in *model training* must be distinguished. During *inference*, the same subsequent token (or a probability distribution over tokens) is predicted as is present at the same position in training sequences which differ only in irrelevant circumstances, wherein these training sequences are encoded in the LLM on which the prediction is based.[26] The homogeneity condition then comes down to the assumption that *token sequences combining a user query with the answer thus far provided by the model determine the subsequent token (or determine at least a somewhat stable probability distribution over a set of possible subsequent tokens)*. In other words, no further circumstances besides the given token sequences should be required for inference, i.e. for predicting the continuation of a token sequence. All other circumstances, e.g. regarding the context in which language is spoken or written, are either irrelevant or their impact is already reflected in the explicitly examined circumstances, i.e. in the tokens of the given sequence. Of course, this assumption is fallible and in certain situations does fail, e.g. when a user input is not concisely formulated. But overall, it seems reasonable to assume that subsequent content is at least probabilistically determined by a well-posed user query and the previous course of conversation between user and LLM.

Similar to the case of inference, homogeneity in *model building* requires that the training sequences are long enough to at least probabilistically determine subsequent tokens. In other words, the circumstances taken into account must be sufficient to reliably identify difference / indifference makers by comparing token sequences in the training set. This seems plausible given that LLMs are trained in terms of blinded token sequences that typically have the context length of the respective model, which is much longer than most user queries (cf. Sec. 4.1). Ultimately, the inferential success of large language models retrospectively justifies both these homogeneity assumptions.

*(8) Necessity and Meaning*

Methods of variational induction like the method of difference are capable of differentiating necessary from spurious or accidental relationships. These methods are usually understood to result in *causal* laws. For example, Mill's influential formulation of the method of difference explicitly uses causal terminology (1843, 455).[27] But variational induction in fact works independently of the type of necessity linking circumstances with a phenomenon, e.g. whether the necessity is causal or definitional (Pietsch 2022, sec. 7.1).

For the relationship between words in word sequences, various kinds of necessity play a role: concerning grammar, the definition of concepts, or factual content regarding e.g. causal relations in the empirical world. As explained, they can all be analyzed in terms of variational induction. With respect to language modeling, this can be seen as a virtue since, for example, empirical and definitional content can often be shifted in the interpretation of natural language without changing the word sequences (cf. debates on the *analytic-synthetic distinction* in the philosophical literature, e.g. Rey 2023).

---

[26] Deriving a prediction from a given model is strictly speaking a deductive step. By contrast, the complete reasoning from the training instances to the prediction can be considered as an inductive inference.

[27] A considerable body of literature exists on whether machine learning algorithms using big data can infer causal relationships (Canali 2016, Pietsch 2016a, Galli 2023, Bujsman 2023, Canali & Ratti 2024, Wunsch et al. 2024). Other work addresses to what extent machine learning algorithms can identify concepts (Boge 2024, Boge & de Regt 2026, Pietsch 2022, Ch. 7).

All these kinds of necessity ensure that meaningful text is produced. Only if the correct difference and indifference makers are identified, does the model generalize reliably to unknown token sequences, i.e. is the model extendable in the sense that unknown token sequences can be meaningfully continued. Learning difference and indifference makers—and thereby identifying necessary relationships—to a certain extent allows for determining the meaning of words and sentences.[28]

In summary, it was shown in Section 3 that the training of large language models as well as predictions based on these models fit the overall structure of variational induction, i.e. an inductive logic identifying difference and indifference makers among the circumstances of a phenomenon. In the next Section 4, I will proceed to argue that certain aspects of large language models not shared by other neural networks address problems specific to the implementation of variational induction for natural language processing.

## 4. Specific Aspects of Large Language Models

As we have seen, large language models have a generic neural network architecture with specific features tailored for human language processing and generation.[29] In the following, I argue that these features fit well with variational induction because they address certain problems arising in the application of a difference making logic to large language modelling. For example, almost always only *sparse data* is available for training LLMs, since even the largest text corpora do not cover all possible utterings in a language. Another crucial issue concerns *long-range dependencies* between tokens in a sequence, which are prevalent in natural language, but cannot be modelled by conventional neural network architectures.

### *4.1 Transformer Architecture*

In the following, an overview of the most important building blocks of large language models is provided. While the building and training of LLMs requires substantial specialized engineering know-how, their underlying architecture is surprisingly simple. LLMs just transform a given input through a series of vector and matrix operations to generate a prediction. This architectural simplicity is particularly striking given that these models are capable of representing human language with impressive sophistication.

[28] In a similar vein, Holger Lyre argues: “A strong argument for [their semantic] grounding is that LLMs form world models, and the evidence for this is that the representational geometry of these models follows semantic similarities.” (2024, 14) LLMs have also been successfully used to predict human behaviour indicating that they can identify causal relationships in the world (Kieval & Buckner 2025).

[29] Stephen Wolfram notes that many aspects of neural network training “have been discovered by trial and error, adding ideas and tricks that have progressively built a significant lore about how to work with neural nets” (Wolfram 2023, 31).

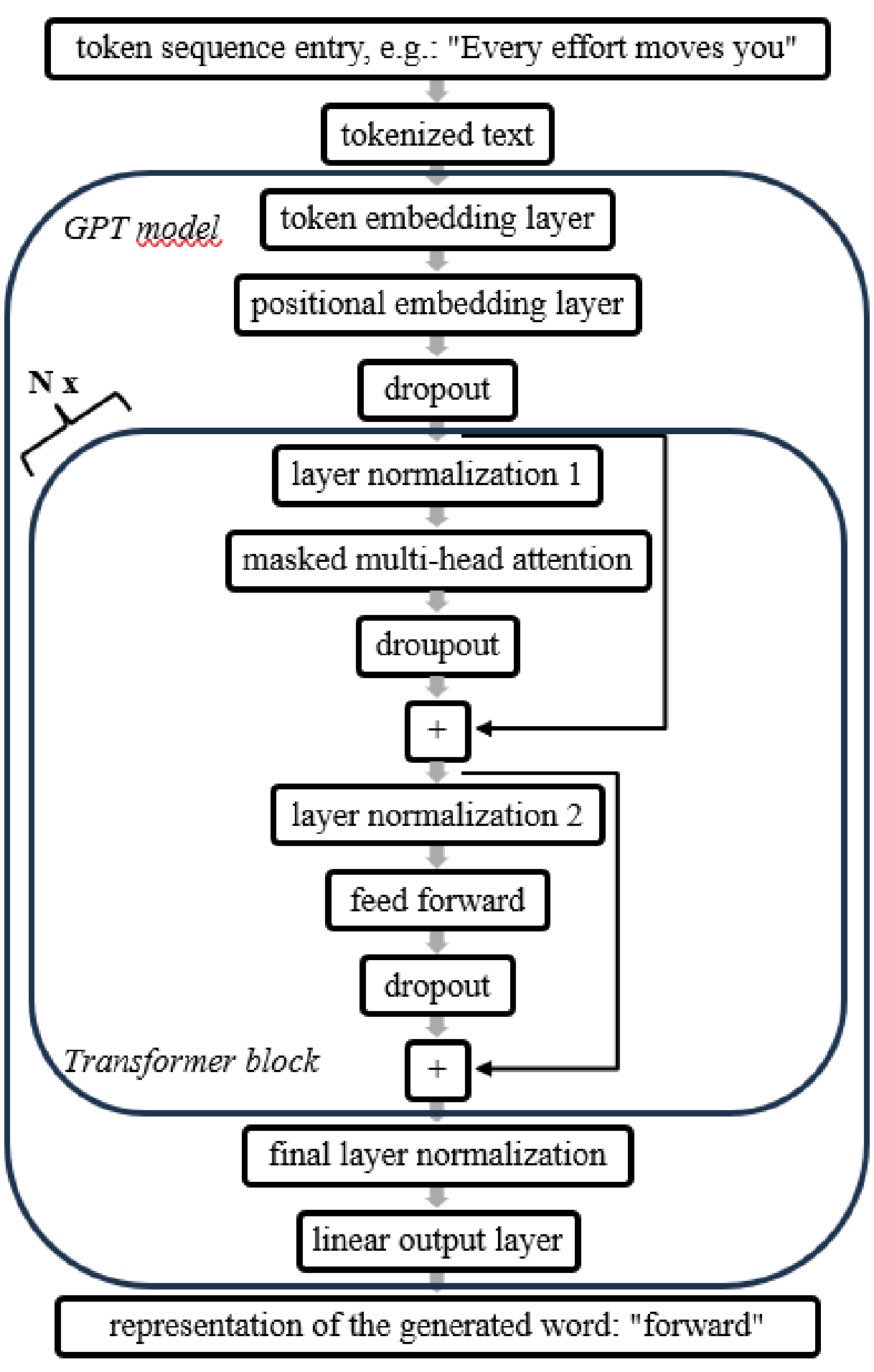

Fig. 6 exemplary architecture of an LLM, here the GPT-2 model (taken from Raschka 2025, 163, 189)

I use OpenAI's GPT-2 model for illustration, as it is the last model architecture that OpenAI fully disclosed, before the success of Chat-GPT led many AI companies to become more secretive (Raschka, Sec. 5.5). The overall architecture of this model is illustrated in Figure 6 below (Raschka 2025, Figs. 4.15, 5.17; Radford et al. 2018, Sec. 3 & Fig. 1; Radford et al. 2019, Sec. 2.3)

First, the text input is tokenized, i.e. converted into numbers. Token and positional embedding layers follow, which change the representation so that the meaning of the text input is more adequately reflected. This concept of embedding will be explained in detail in Section 4.2 below.

The subsequent transformer block, as originally proposed in Vaswani et al. (2017), is repeated *N* times. It comprises a multi-head attention module as discussed in Section 4.4 and a rather conventional feed-forward neural network module. When passing through the transformer block, “the output is a context vector that encapsulates information from the entire input sequence” (Raschka 2025, 116).

The multi-head attention module analyzes relationships between tokens regardless of their distance in the input sequence (Raschka 2025, 113). The feed-forward network consists of two linear transformations with a non-linear activation in between, temporarily expanding the hidden dimension before projecting it back to the original size. The expanded dimension allows for a richer representation space which enhances the model’s abilities to learn complex relationships (Raschka 2025, Sec. 4.3, Fig. 4.9).

The final linear layer of the model projects the output sequence of the last transformer block onto a sequence of vectors, each having the dimension of the vocabulary size. These vectors are transformed into probability distributions by applying the so-called softmax function (Raschka 2025, 123). The softmax function, and thus the resulting probabilities, can be adjusted using a temperature parameter as discussed in Section 4.3. During model training, all output vectors are utilized to calculate the loss across the entire sequence. In contrast, for inference, only the vector representing the predicted next token is used to extend the given input sequence.

The further elements depicted in Figure 6—dropout, shortcut connections, and normalization—are all standard operations in neural network modeling. They speed up convergence, stabilize the training process and serve as a countermeasure against overfitting (Raschka 2025, Secs. 3.5.2, 4.2, 4.4).

LLM architectures are characterized by several meta- or hyperparameters (cf. Raschka 2025, 95) including: the vocabulary size used for tokenization (50257 for the largest GPT-2 model); the embedding dimension, i.e. the dimension of the vectors into which the individual input tokens are transformed (1600 for the largest GPT-2 model, 12288 for the largest GPT-3 model[30]); the context length denoting the maximum number of input tokens which a model can handle (1024 for GPT-2, 2048 for GPT-3[31]); the number of transformer blocks in the model (48 for the largest GPT-2 model, 96 for the largest GPT-3 model); the number of heads in multi-head attention (25 for the largest GPT2-model, 96 for the largest GPT-3 model). As a result of these architectural changes, the number of trainable parameters is, for example, 1.558 billion for the largest GPT-2 model and 175 billion for the largest GPT-3 model (cf. Raschka 2025, 163).

In summary, LLMs employ a neural network architecture that does not rely on any assumptions regarding syntax or semantics of the modeled language. Instead, syntax and semantics are encoded completely in the trainable weights learned from data. Remarkably, across successive generations of GPT models, the underlying architecture has barely changed (Raschka 2025, 94). Instead, the above-mentioned parameters were systematically scaled up, leading to substantial improvements in performance. This reflects OpenAI's big and ultimately successful bet that the necessary structural elements were already there in the early models and primarily needed to be scaled.

[30] The numbers for the GPT-2 and GPT-3 models are taken from Wolfram (2023, 47-54), Radford et al. (2019) and Brown et al. (2020), see also Raschka (2025, 90).

[31] More recent GPT models have much larger context lengths of 100,000 tokens and more (see openai.com/index/new-models-and-developer-products-announced-at-devday/ accessed on 5.1.2026)

### *4.2 Token Embeddings and Semantics*

In order to process human language on a computer, it must first be converted into numbers that can be fed into a neural network as numerical input. As already mentioned, for practical reasons and for greater flexibility, e.g. to handle new word creations, these numbers are assigned to so-called tokens, which can be whole words, but often are smaller units such as word fragments or punctuation marks. As Stephen Wolfram notes, various approaches to language modeling like word2vec, GloVe, BERT or GPT are all based on a different neural net approach, but "ultimately all of them take words and characterize them by lists of hundreds to thousands of numbers" (2023, 45).

A simple way to turn a word or, more precisely, a token into numbers is by means of a *one-hot encoding* given by a vector, which contains a value of '1' at the position corresponding to the represented token and '0' elsewhere (cf. Sec. 2.1). Obviously, the dimension of these vectors must equal the size of the vocabulary, i.e., the number of possible tokens. The primary drawbacks of the primitive one-hot representation are twofold: first, it does not encode semantic information, i.e. semantic relatedness is not reflected in geometric proximity between vectors, and second, the number of dimensions—and therefore the computational cost—is unnecessarily high.[32]

These issues can be addressed by training a neural network to transform the one-hot encoding of tokens into a more compact representation: an embedding. Embeddings capture semantic relationships, for example, by positioning synonyms close to each other in vector space and by mapping analogies through similar geometrical relations. In practice, the one-hot encoding is typically represented by an index or ID number that identifies the position of the non-zero entry in the corresponding one-hot vector (see Fig. 2). Similarity relationships between tokens are then represented by a trainable embedding matrix, which corresponds to a simple look-up table that maps each token index to an embedding vector. Accordingly, the dimension of this look-up matrix is vocabulary size times embedding dimension.

The embedding dimension is much smaller than the vocabulary size, which is crucial for information reduction (cf. item (6) in Sec. 3.3). The effectiveness of this compression reflects the fact that many different syntactical instantiations, which constitute the input to an LLM,

---

[32] https://developers.google.com/machine-learning/crash-course/embeddings/embedding-space (accessed on 10.10.2025)

can express the same underlying semantics, as there are always many largely equivalent ways to convey the same idea in words. Embeddings encode abstract linguistic regularities that are shared across contexts and even across languages. This abstraction is a key reason why such models also perform well at tasks like translation between natural languages or between programming languages.

In natural language processing by LLMs, sequences of tokens in the text corpora used for training are the *only* type of information, from which semantic relationships are inferred (cf. Sec. 2.2). While this assumption could be criticized as being overly reductionist, and it is not initially plausible that semantic content can be reduced to the relationship between a word and its neighbors in sequences of words, the success of large language models in hindsight justifies this underlying approach. For example, that the words 'purchase' and 'buy' are synonyms can be concluded from the fact that these words can be replaced by each other in token sequences. Accordingly, there are examples in the training corpora, where these synonyms are combined with the same or similar neighboring tokens. For example, a corpus might include the phrases: "my mom purchased a car" and "his aunt bought a VW Passat".

However, the token vectors of the embedding layer lack information about the positions of the tokens in the original input sequence. Also, the self-attention mechanism to be discussed in Section 4.4 does not distinguish token positions. Therefore, positional information is typically injected into the network by adding positional vectors to the embedding vectors. These positional vectors have the same dimension as the embedding vectors and are unique to each position in the input sequence, ensuring the model can account for word order (Raschka 2025, Sec. 2.8).

From the perspective of variational induction, embeddings play an important role in addressing the problem of *sparse data*. No matter how large the training corpora are, they will almost never include examples that are syntactically similar enough to a previously unknown token sequence for which the next token is to be predicted. Therefore, changing to a more compressed semantic representation is essential. Based on an embedding and the semantic similarity-relations encoded in the embedding, an LLM can effectively search the training corpora for word sequences most similar *in meaning* and base the prediction on these related word sequences. In a way, embeddings reparametrize the high-dimensional landscape, in which optimization procedures like gradient descent operate, so that model parameters representing semantically similar concepts are geometrically related to each other, e.g. spatially close or in spatially analogous vectorial relationships. The success of LLMs in generating meaningful content shows that the training corpora contain sufficient information for the model to reliably identify difference and indifference makers in the compressed embedding representation.

### *4.3 Temperature Parameter and Determinism*

In conventional model building, statistical models are usually employed when training data is insufficient to derive a deterministic model. Given the complexity and context-dependency of language, it is unrealistic to expect deterministic models, which predict the subsequent token with certainty. As explained in Section 2.1, LLMs are statistical in that they yield as outcome a probability vector with a dimension equal to the vocabulary size (see Fig. 2). This vector assigns a probability to each token $j$ in the vocabulary that this token continues the given token sequence. These probabilities are obtained by applying the *softmax function* to the output layer of the LLM: $\text{softmax}(z_i, T) = \frac{e^{z_i/T}}{\sum_{j=1}^{K} e^{z_j/T}}$. The result is a normalized vector with entries between zero and one that sum to one. Here, the $z_j$ denote the values—so-called *logits*—of the individual tokens j in the output layer and $K$ is the vocabulary size.

The probabilities can be tuned using the parameter $T$—called "temperature" due to a formal analogy with statistical mechanics. This temperature parameter is typically used for inference, but not during training. Lower temperatures result in a sharper distribution with either one or a few highly probable tokens and higher temperatures in a more uniform distribution. It turns out that token sequences generated by LLMs are more interesting, more diverse, and more similar to actual human language, if not always the most probable token or token with the highest rank is selected (Raschka 2025, Sec. 5.3.1; Wolfram 2023, p. 2-6). Conversely, restricting probabilistic sampling to a smaller number of the most probable tokens excludes those with very low probability that may lead to nonsensical continuations (Raschka 2025, Sec. 5.3.2).

Although the basic framework of variational induction is deterministic, it can be generalized to cover probabilistic inferences. As explained under item (5) of Section 3.3, difference and indifference making then regards probability distributions over several phenomena rather than a single phenomenon. Such a statistical generalization of variational induction is useful, either when a phenomenon is not fully determined by its circumstances—as is presumably the case in quantum mechanics—or when some relevant circumstances are not taken into account for the prediction of a phenomenon. In the case of LLMs, it is plausible that there are relevant circumstances not covered by the token sequence given by query and answer provided so far: e.g. circumstances regarding the psychology of a speaker or regarding the real-world context, in which language is uttered. Furthermore, the probabilistic nature may to some extent account for creative aspects of language. Thus, the perspective of variational induction corroborates that the output of an LLM should be a probability distribution of tokens, rather than a single token.

### *4.4 Attention Mechanism and Long-Range Dependencies*

In the past, recurrent neural networks (RNNs) were widely used for modeling sequential data such as human language. In RNNs, information must be propagated successively through a chain of hidden states so that the number of processing steps between two tokens grows with their distance (e.g. Vaswani et al. 2017, sec. 2; Raschka 2025, 52-54). Therefore, these architectures were poorly suited for capturing *long-range dependencies in token sequences.* However, such dependencies are quite common in natural language, where meaningful relationships often span long stretches of text—for example, when earlier content is referenced much later.

A major breakthrough in the development of large language models came with the introduction of a fundamentally different modelling approach that addressed the problem of long-range dependencies. This new architecture was proposed in the seminal paper "Attention Is All You Need" (2017) by Google researchers Vaswani et al., who argued that language models could be based entirely on the so-called *attention mechanism*, dispensing with recurrent components altogether. A particular neural network architecture based on the attention mechanism is known as the *transformer*, the 'T' in GPT: "In the Transformer [the number of operations required to relate signals from two arbitrary input or output position] is reduced to a constant number of operations, albeit at the cost of reduced effective resolution due to averaging attention-weighted positions, an effect we counteract with Multi-Head Attention." (Vaswani et al. 2017, Sec. 2)

The central concept is *self-attention*, which provides a measure of how strongly a certain token relates to all other tokens in a sequence. This relatedness is determined by multiplying the embedding vectors of the tokens in the input sequence with three different weight matrices, known as query, key and value matrix. The terminology reflects the fact that the attention mechanism was originally conceived by comparison with a simple database query, although in hindsight the analogy seems limited (Raschka 2025, 70). The three matrices are learned during training, i.e. the entries of the matrices are updated as part of the model optimization process. This is crucial for enabling large language models to learn how to adequately capture contextual information for the individual tokens in a sequence. The relationship between a token and other tokens in a sequence is encoded in *context vectors* discussed in the following (Raschka 2025, 65).

Multiplication with the matrices yields three distinct vectors for each embedding vector of a token: a query vector, a key vector, and a value vector. For each token $i$ in the input sequence, the dot product, i.e. element-wise multiplication and subsequent summation, of its query vector with the key vectors of all tokens $j$ in the sequence is calculated—including the dot-product with the key vector of token $i$ itself. The resulting numbers are scaled, normalized by applying the softmax function, and then multiplied with the respective value vector. The resulting vectors for all tokens $j$ are summed to yield the context vector for token $i$. The procedure is repeated for every token $i$ in the input sequence (Raschka 2025, 65-70). The approach is called 'self-attention' because the relationships between tokens in the same sequence are examined rather than the relationships between two different sequences as for example in a translation task.

Let me briefly point out two further aspects in connection with the attention mechanism: multi-head attention and the stacking of multiple transformer blocks (see Fig. 6). In a multi-head attention setup, several attention mechanisms are executed in parallel, each with its own weight matrices that are updated independently of each other during training. The resulting context vectors from the individual heads are concatenated to form a context matrix. This approach allows the model to simultaneously encode diverse aspects of context—such as definitions, grammatical dependencies, or empirical facts.

In addition, several transformer blocks each comprising a multi-head attention module and a feed-forward module are stacked sequentially. By analogy with other neural network architectures, this hierarchy of subsequent transformer blocks likely allows the model to resolve linguistic features at various levels of abstraction and detail. The nodes in hidden layers of the LLM presumably represent intermediate conceptualizations (Raschka 2025, 82; cf. also item (4) in Sec. 3.3 above).

From the perspective of variational induction, the transformer architecture with its attention mechanism allows for identifying difference and indifference making relationships even at long distances between tokens. Also, various kinds of difference making, concerning for example grammatical, definitional or empirical aspects, can be analysed at different levels of resolution. In conventional variational induction, such hierarchical structures are usually represented in terms of Boolean combinations of circumstances. Large language models generalize the Boolean approach by employing a multi-layer architecture determined by trainable continuous parameters, which is well-suited to model gradual dependencies and allows for probabilistic predictions.

## 5. Conclusion

Natural language is highly complex and almost certainly irreducible to a simple model. However, large language models with billions of trainable parameters and trained with vast digitized text corpora seem capable of representing this complexity, while the underlying maths is surprisingly simple consisting of only a large number of vector and matrix operations.

Importantly, large language models do not merely interpolate between data but rather they learn difference- and indifference-making relationships between tokens in a sequence. These relationships generalize well to unknown token sequences, which is why large language models generate meaningful word sequences on the level of human written and spoken language. As shown, the difference- and indifference-making relationships are learned in terms of variational induction, essentially through the methods of difference and agreement, which are implemented when the model is improved during training.

The overall result is not surprising, since, in the history of scientific method, variational induction has in various disguises been the only inductive method that under the right premises reliably works. Examples are: the experimental method itself, where circumstances of a phenomenon are systematically varied to examine their impact on a phenomenon; randomized control trials, where control and study groups are statistically defined to be compared in terms of difference making; or, more recently, decision trees in machine learning, which proceed by iteratively determining the most important difference makers amongst a number of circumstances. Thus, the logic behind LLMs is not new, it is essentially a variant of the experimental method; however, it is applied in novel ways using the enormous computational resources that have become available.